# Clustering Time-Series by a Novel Slope-Based Similarity Measure Considering Particle Swarm Optimization

Hossein Kamalzadeh [a1], Abbas Ahmadi [b], Saeed Mansour [c]

[a] Department of Engineering, Management, Information and Systems, Southern Methodist University, Dallas, TX, US, email: hkamalzadeh@smu.edu
[b] Department of Industrial Engineering & Management Systems, Amirkabir University of Technology, Tehran, Iran, email: abbas.ahmadi@aut.ac.ir
[c] Department of Industrial Engineering & Management Systems, Amirkabir University of Technology, Tehran, Iran, email: s.mansour@aut.ac.ir

## Abstract

Recently there has been an increase in the studies on time-series data mining specifically time-series clustering due to the vast existence of time-series in various domains. The large volume of data in the form of time-series makes it necessary to employ various techniques such as clustering to understand the data and to extract information and hidden patterns. In the field of clustering specifically, time-series clustering, the most important aspects are the similarity measure used and the algorithm employed to conduct the clustering. In this paper, a new similarity measure for time-series clustering is developed based on a combination of a simple representation of time-series, slope of each segment of time-series, Euclidean distance and the so-called dynamic time warping. It is proved in this paper that the proposed distance measure is metric and thus indexing can be applied. For the task of clustering, the Particle Swarm Optimization algorithm is employed. The proposed similarity measure is compared to three existing measures in terms of various criteria used for the evaluation of clustering algorithms. The results indicate that the proposed similarity measure outperforms the rest in almost every dataset used in this paper.

## 1. Introduction

Nowadays time series data are being produced in many areas including medical and health care, scientific, financial, economic, governmental, industrial, environmental and socioeconomic phenomena, and these phenomena's attributes are not always static and are usually changing over the time [1], [2], [3], [4], [5], [6], [7], [8], [9], [10]. There are different kinds of time series data including univariate or multivariate, discrete or real-valued, uniformly or non-uniformly sampled and time series of equal or unequal length [11]. Therefore, the exploration and analysis of these data are essential and of high interest in order to find useful information and patterns hidden inside them [2]. Recently the amount of this data and its dimensionality has increased with such a great speed that rarely the traditional methods of analyzing data can deal with [1], [4], [12]. Thus, different techniques with different purposes have been developed to analyze time series data, within the framework of classification, clustering, prediction and forecasting, outlier detection and noise removal [13].

Among the mentioned techniques, clustering, which is proved useful in the field of data mining, especially when the data is big, is to organize the data into several groups, whose items have the most similarity to each other and are as much as possible different from the other cluster's data [3], [4], [6], [14]. This way there will be clusters in which data's characteristics are very similar to other data's of the same cluster and therefore the understanding of the whole data would be much easier [5] [15].

---

[1] Corresponding author: Department of EMIS, Lyle School of Engineering, Southern Methodist University, P.O. Box 750123, Dallas, TX 75275-0123, email: hkamalzadeh@smu.edu, Tel: +1 (213) 292 1746

Clustering methods for static data have been divided into five groups including partitioning methods, hierarchical methods, grid-based methods, density-based methods, and model-based methods; For dynamic data, as they are more complex, there are some other classifications for the methods including raw-data based methods, feature-based methods and model-based methods [3], [10]. It has also been categorized into two main categories: whole clustering and subsequence clustering [4], [10], [16] [17]. Therefore, many algorithms have been used to conduct the clustering task of time series data, each dealing with a specific type of time series and in different applications. However, according to [18], most of these algorithms somewhat modifies either the time series data or the algorithm used for static data in order to make them compatible to be used with each other. A more comprehensive study of time-series clustering can be found in the book "Time-Series Clustering and Classification" recently published in 2019 [19].

Although there are many different terms and characteristics related to the field of clustering, most of the studies in literature have focused on two major properties of cluster analysis especially when dealing with time series data, including the main algorithm used for clustering time series and the similarity measure which is used to calculate the distance between the time series.

The algorithms used in clustering can be divided into two major groups which are evolutionary and non-evolutionary algorithms. Some non-evolutionary algorithms are k-means [3] [5] [8], I-k-means [6], k-harmonic means [20], fuzzy clustering [9], hybrid fuzzy c-means and fuzzy c-medoids [1], kernel k-means [11], and some hierarchical methods [4], [7], [12]. These non-evolutionary algorithms usually provide poor results because they are dependent on the initial solution and they lack power and robustness when dealing with high dimensional data [15]. Contrary to this group of algorithms, the evolutionary ones have been proved to be more useful and robust and not having the limitations of the former ones [21]. These methods are inspired by the collective intelligence and the intelligent behavior of a swarm or flock of birds, ants or bees existing in nature and they are divided into two main groups that are ant-based clustering and Particle Swarm Optimization-based clustering [15], [21], [22], [23], [24], [25], [26], [27]. There are also some studies on time-series clustering that do not belong to or cannot be categorized into the mentioned categories such as network-based approaches [28], covariance-based clustering [29], fuzzy clustering [30], and Reinforcement techniques such as Hidden Markov models [31] or Partially Observable Markov Decision Processes [32,33].

The second important property of the clustering methods is the distance measure used for calculating the distance between two data points. This measure is also called the similarity measure as the similarity between the patterns inside the data can be recognized with this measure [15]. There are many different similarity measures used with static data that can be modified to dynamic data too [34]. Some of these measures which are mostly used with static data are Minkowski distance and Euclidean distance [15], [21], [22], [23], Mahalanobis distance [15], kernel-based similarity measure [25] and Cosine similarity [15]. Because the time series data are more complex than the static data, these measures are often weak in dealing with dynamic data and therefore some other similarity and distance measures have been developed lately to deal with time series data [34], [35]. These measures are Pearson's correlation coefficient [12], auto-regressive models [13], dynamic time warping [1], [4], [36], [37], [38], [39], [40], [41], [42], [43], edit distance [34], [44], time-warped edit distance [45], minimum jump costs dissimilarity [34], PCA-based distance [46], [47], cross-correlation measure [48] [49], and Euclidean distance [6], [11], [17].

According to many literature reviews in this field, time-series clustering specifically whole time-series clustering could be divided into four main components including time-series representation, similarity or distance measure, clustering prototypes, and time-series clustering algorithm [10],[14], [16], [18]. The accordance of these four components are very important and the technique selected for each component should match the rest of the techniques. For example, many distance measures can't be used while a specific representation method is employed to reduce the size of the time-series [50]. Distance measure affects the quality of clustering to a great extent, just like cluster prototypes which in some cases are as important as the distance measure [10]. These facts indicate that to have the best results for the clustering of a given dataset of time-series, each component must be taken into consideration and appropriate techniques for each one should be selected [50], [51]. Thus, in this paper, since the most important component is considered to be the distance measure, the main contribution of this paper is that a new similarity (distance) measure is developed specifically for time-series data. Since there is a great relationship between the data representation and the distance measure used for clustering, the distance measure is developed based on the representation form provided by the algorithm developed in [52]. Thus the representation form is indisputably selected from [52]. The novelty of the proposed similarity measure is that it takes into account a combination of a simple representation of time-series, slope of each segment of time-series, and Euclidean distance all at the same time. In other words, the proposed similarity measure takes the actual physics of the time-series into account. Finally, PSO is selected to perform the clustering as it is proved to be very efficient in this job.

The rest of the paper is organized as follows. In section 2 a brief description of time series and some definitions related to clustering are given for a better understanding of the whole problem. Section 3 presents the proposed distance measure for time-series clustering. Then in section 4 Particle Swarm Optimization and more specifically PSO clustering are discussed briefly. In section 5 experimental results for both the PSO-based clustering and the proposed distance measure compared to others of their kind are provided. Finally, section 6 concludes the paper. The proof for the proposition made in section 3 is provided in the appendix.

## 2. Preliminaries and Definitions

To better understand the whole problem and have a common language while talking about clustering the time-series data, it is better to provide some definitions regarding time-series and clustering.

### 2.1. Time-series

**Definition 1**: A time-series $TS$, is a sequence of sampled values of data which are put in order chronologically. The length of a time series is defined by the number of time steps during which it has been sampled. Therefore, a time-series with $N$ time steps is shown by $TS = \{x_1, x_2, ..., x_i, ..., x_N\}$. When the time series is univariate, each $x_i$ is a single value and therefore the size of the time-series is $1 \times N$. But when it is a multivariate time-series, each $x_i$ includes a number of values according to the number of features or variables ($m$) sampled at a specific time step. Therefore, the size of a multivariate time-series is $m \times N$ [17].

When dealing with a data set consisting of $T$ different time-series, the size of the input data is $T \times m \times N$. Note that in this paper $TS_i^{(j)}$ is the $i^{th}$ point from the Time-series ($j$) in the database.

### 2.2. Time-Series Representation

Since time-series specifically the multivariate ones are of high dimension and processing them needs a lot of memory, various techniques have been developed to reduce their dimension without losing any important characteristics [10].

**Definition 2**: Time-series representation or dimensionality reduction is to transform a given raw time-series $RTS = \{u_1, u_2, ..., u_i, ..., u_{N'}\}$ into a lower dimension space $TS = \{x_1, ..., x_N\}$ where $N < N'$ by reducing its data points or feature extraction.

Selecting an appropriate representation method can greatly improve the clustering quality as well as reducing the time and memory requirement for the task of clustering. Representation methods are divided into four categories including data adaptive, non-data adaptive, model-based and data dictated. In this paper, the technique called APSOS (Adaptive Particle Swarm Optimization Segmentation) proposed in a former paper from the authors [52] is employed to reduce the dimension of data. APSOS is a method of time-series segmentation and is somehow adaptive to each time-series and tries to reduce its dimension (time steps) while preserving its shape as much as possible. In this approach, a limited number of the time-series' points are selected to reconstruct it. The goal is to select the optimal points in order to reduce the error. It is proved in [52] that this technique outperforms most of the techniques used in this area.

### 2.3. Similarity or Distance Measure

Following what was mentioned about the distance measures in the previous section, important factors should be considered in selecting an appropriate distance measure for the task of time-series clustering. These factors include the level of analyzing (shape level or structure level) and the objective of analyzing (to find similarity in time, in shape or in change), which as a result define the type of distance measure that should be used [10]. When the objective is to find similar time-series in shape, i.e. the occurrence time of similar patterns is not important, time elastic measures such as dynamic time warping [53] become useful [10]. Since the output of the representation method described above is usually a very short time-series and also is based on the time-series' shape, a shape-based distance measure should be employed for the clustering task.

## 2.4. Clustering

Clustering is to organize and put a set of data samples consisting of $M$ items into $K$ clusters in a way that the intra-cluster distance is minimized and/or the inter-cluster distance is maximized. Clustering could be applied to both static and dynamic data, but each group needs its specific methods. Many techniques have been proposed for clustering of static data [10], which also could be used with dynamic data by applying some modifications. As it was mentioned in the previous section, these techniques can be classified into five major groups including partitioning, hierarchical, grid-based, model-based, and density-based.

Mainly there are two groups of measurement with which the accuracy of a clustering solution can be evaluated. The first group is dependent on external information about the data such as supplied labels. In the absence of such information, internal criteria could be useful. External validity measures which are based on the amount of agreement between the externally provided truth and the clustering solution are as follows:

**Cluster Purity**: if the solution to a clustering algorithm is $C = \{C_1, C_2, ..., C_K\}$, then each cluster is assigned to one of the known classes which is the most abundant one in the cluster. Then the ratio of the number of correctly assigned objects to the number of all objects is called the cluster purity [10].

**Cluster Similarity Measure (CSM)**: the basis of the CSM is similar to the purity measure. Let $G$ and $C$ respectively denote the ground truth of $K$ classes and the clustering solution with $K$ clusters. Then the CSM would be [18]:

$$CSM(G, C) = \frac{1}{K} \sum_{i=1}^{K} \max_{1 \leq j \leq K} Sim(G_i, C_j), \quad (1)$$

where $G_i$ is the $i^{th}$ class labeled externally and $C_j$ is the $j^{th}$ cluster of the obtained solution and:

$$Sim(G_i, C_j) = \frac{2 \times card(G_i \cap C_j)}{card(G_i) + card(C_j)}, \quad (2)$$

where $card(.)$ is the number of the members of the set denoted inside it.

**Jaccard Score**: this measure is defined as follows [54]:

$$Jaccard = \frac{a}{a + b + c}, \quad (3)$$

where

- $a$: number of pairs having the same class label in $G$ and clustered in the same cluster in $C$.
- $b$: number of pairs having the same class label in $G$ but clustered in different clusters in $C$.
- $c$: number of pairs with different class labels in $G$ but clustered in the same cluster in $C$.
- $d$: number of pairs with different class labels in $G$ and clustered in different clusters in $C$.

**Rand Index (RI)**: this measure is calculated as follows [54]:

$$RI = \frac{a + d}{a + b + c + d}. \quad (4)$$

**Folkes and Mallow index (FM)**: this index is given below [54]:

$$FM = \sqrt{\frac{a}{a + b} \times \frac{a}{a + c}}. \quad (5)$$

In the absence of ground truth there are some internal cluster validity measures to evaluate the solution of a clustering algorithm, also called fitness functions. Some of these quality measures of clustering techniques are given next [10], [15], [21].

**Compactness Measure:** also called within-cluster distance, shows how similar are the members of each cluster. This measure is defined as

$$F_c(M) = \frac{1}{K}\sum_{k=1}^{K}\frac{1}{n_k}\sum_{j=1}^{n_k}d(m^k, y_j^k), \qquad (6)$$

Where $M = (m^1, m^2, \ldots, m^K)$, $m^k$ is the center of the cluster $k$, $K$ is the number of clusters, $n_k$ is the number of samples in cluster $k$, $y_j^k$ is the $j^{th}$ member of the cluster $k$ and $d(.)$ is the distance between two samples. The goal of the clustering algorithm is to minimize this measure [15], [21].

**Separation Measure:** also called inter-cluster distance, shows the separation between clusters. It is defined as

$$F_s(M) = \frac{1}{K(K-1)}\sum_{j=1}^{K}\sum_{k=j+1}^{K}d(m^j, m^k). \qquad (7)$$

The goal is to maximize this measure [15], [21].

**Combined Measure:** this validity measure is a combination of the last two measures and it simultaneously measures both the compactness and separation of the obtained solution. It is defined as

$$F_{Combined}(M) = w_1 F_c(M) - w_2 F_s(M), \qquad (8)$$

where the weight parameters $w_1$ and $w_2$ are chosen in a way that $w_1 + w_2 = 1$ [15], [21]. This measure is also called weighted inter-intra index [10].

**Sum of Squared Error (SSE):** the so-called SSE could also be used to evaluate the accuracy of a clustering solution [10]. This way, for each time-series of data set, the error is its distance to the nearest cluster.

## 2.5. Cluster Prototype

What makes clustering specifically in the case of time-series a complicated task is how to define the clusters' prototypes or representatives. In static cases, finding the cluster center is much easier since the number of attributes for all data points is equal. In dynamic ones, when it comes to the cases in which the time-series lengths are not equal, finding the cluster center becomes difficult and cannot be accomplished by the traditional averaging methods. Since the quality of clustering depends a lot on the way the cluster prototypes are defined, researchers have employed different approaches for defining the cluster centers in time-series clustering [10]:

**Medoid sequence of the cluster:** in order to find this sequence, the distance between each pair of a cluster is calculated and the one with the lowest SSE is selected as the medoid or center of the cluster.

**Average sequence of the cluster:** this method is similar to the one used for static data and the prototype here is simply a sequence consisting of the mean of all the time-series at each point. But this method could only be used properly when the time-series are from equal length.

**Local search prototype:** in this method first the medoid is found and then the prototype is calculated using the average sequence approach.

## 3. The Proposed Distance Measure: Bilateral Slope-Based Distance

Based on the previous section, to evaluate the quality of a clustering solution, a fitness function such as the combined measure must be calculated, through which the distance between a data instance and a cluster center is calculated. Also to assign a data instance to the nearest cluster, the distance between the data instance and the cluster centers must be calculated. Both of these tasks need a distance measure that is able to correctly calculate the distance between a data instance and a cluster center. Since this paper deals with time-series data and this kind of data is more complex than the static one, an appropriate distance measure is needed. According to the literature, many distance measures have been developed for both static and dynamic data, some could be used with both and some could only be used with one group [10], [18], [34], [55], [56], [57], [58]. The important factor while dealing with time-series data is that the distance measure selected should definitely be compatible with the representation form of the time-series, the algorithm used for clustering and the way the cluster prototypes are defined. Since the cluster prototype approach is selected based on the distance measure and the algorithm is usually modified based on the selected distance measure, the only factor in selecting an appropriate distance measure is to know how the time-series are going to be dealt with and how they are represented and whether they are going to be analyzed directly in the raw form or a dimensionality reduction technique has been applied on them. It is believed that the time-series' representation greatly affects the type of distance measure which is going to be used for the clustering. Since time-series are usually from high dimensions, their dimension would be reduced by one of the existing methods for dimensionality reduction and as APSOS has been proved to perform the job more efficiently than other methods of segmentation [52], in this paper the distance measure is developed based on the output of APSOS.

APSOS is a segmentation algorithm which selects a small but sufficient number of points from the raw time-series to build a new time-series very similar to the original one but in a much lower dimension. APSOS tries to keep the original time-series' shape and visual characteristics such as local extremums by keeping the most important points from the time-series. APSOS is based on PSO with the goal of reducing the amount of reconstruction error through iterations. According to [52] the outcome of the algorithm is highly similar to the original time-series and therefore could be used instead with the advantage of having much less number of time steps. Figure 1(a) simply shows a raw time-series with 106 time-steps and its segmented form with only 5 time-steps produced by APSOS. According to this figure, the APSOS selects the points that best represent the original time-series' shape. Simple form of the APSOS outcome is a sequence of values which indicates the transformed time-series. For the example, in Figure 1(a) this sequence is shown in Figure 1(b) as a new time-series. The new time-series seems to have the original one's shape but the slopes seem to be different. Putting them together in one frame, Figure 1(c) shows the difference more clearly. According to Figure 1(c) APSOS preserves the original time-series' shape but can't save the rate at which the time-series has been sampled. APSOS doesn't preserve the sample rate and it alters it by not saving the segments' slopes. The fact that APSOS alters the sample rate is not a deficiency of this method but is a limitation for it, making it only compatible with elastic distance measures such as Dynamic Time Warping; distance measures on which different sample rates have no negative effect. Thus the distance measure which is going to be developed should be elastic. Here one of the best, Dynamic Time Warping, is selected as the basis but modified to best adapt with the APSOS output.

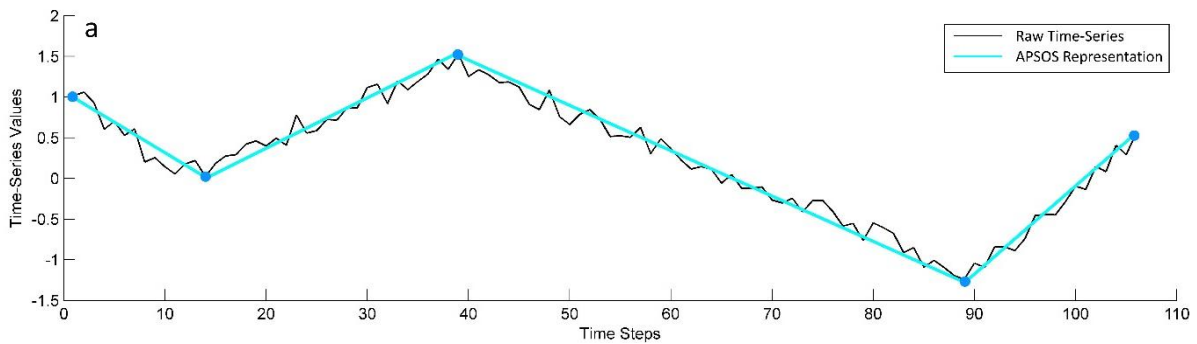

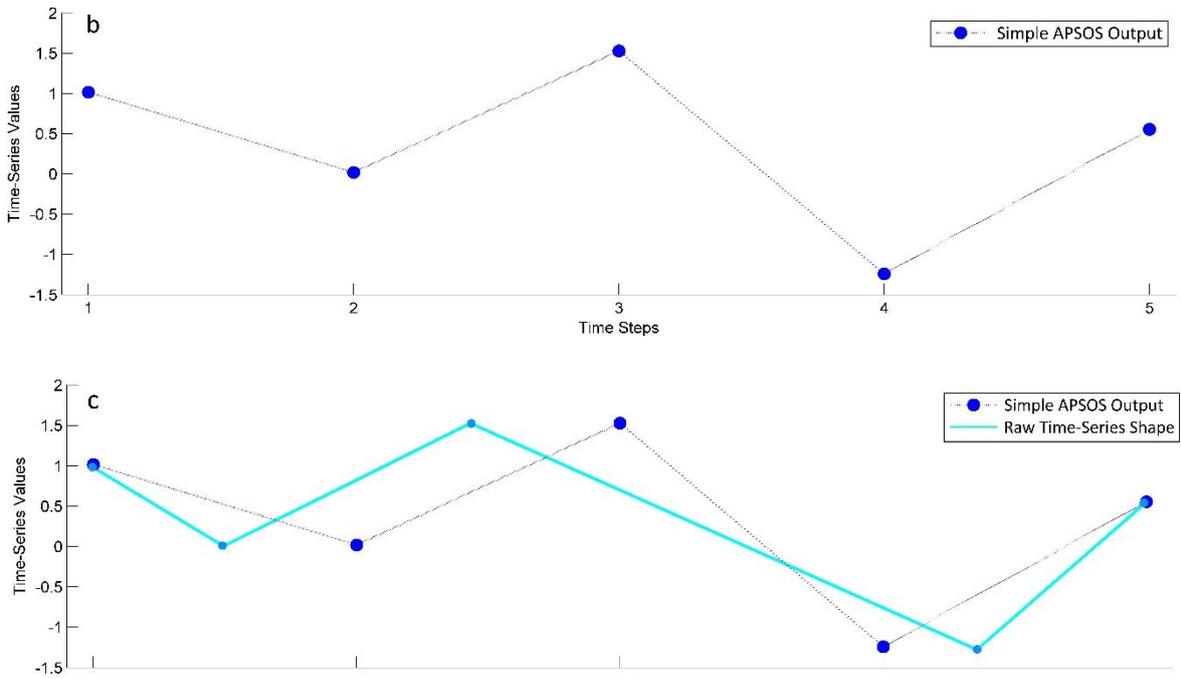

*Figure 1 (a) A raw time-series and its APSOS representation, (b) the sequence representing the APSOS output, (c) sample rate alteration and slope variances*

What is important here is that the slope of each segment must be considered in the APSOS output. With a small modification of the algorithm, the outcome can also contain the slope of each time-series' segment. Consider $RTS = \{u_1, u_2, \ldots, u_i, \ldots, u_{N'}\}$ as the raw time-series and $TS = \{x_1, \ldots, x_N\}$ as the APSOS output. By considering the slopes in the form of the sine of the angle formed by the segment line and the horizontal line parallel to the timeline, the APSOS output will be $TS = \{(x_1, \sin\theta_1), \ldots, (x_{N-1}, \sin\theta_{N-1}), (x_N)\}$. Figure 2 shows how $\theta_i$ are defined.

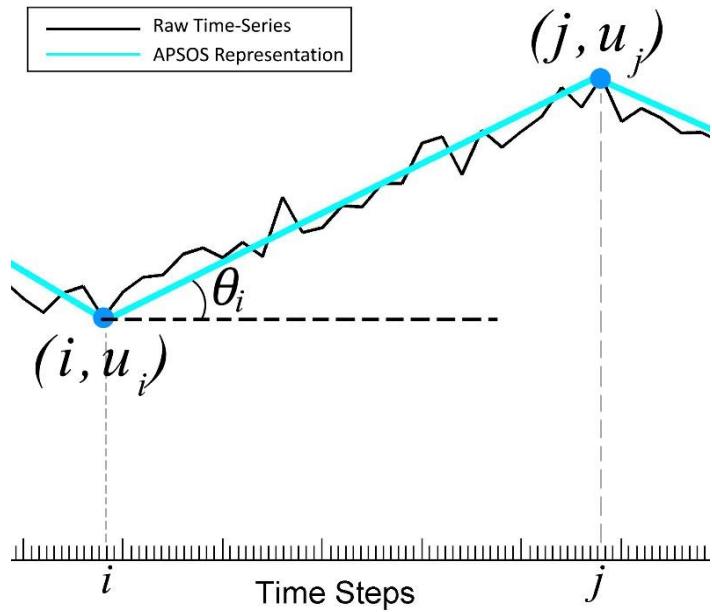

*Figure 2 The θ angle of the i<sup>th</sup> segment of the time-series*

Based on Figure 2 the angle $\theta_i$ would be defined and calculated by

$$\theta_i = Arc\tan\left(\frac{u_j - u_i}{j - i}\right) \ni -\frac{\pi}{2} < \theta_i < \frac{\pi}{2}. \tag{9}$$

Thus $\theta_i$ will always be between $-\frac{\pi}{2}$ and $\frac{\pi}{2}$. Now APSOS can preserve the shape of the time-series more accurately by providing the slope of each segment.

The main goal of this section of the paper is to develop a new distance measure based on the new APSOS outcome and Dynamic Time Warping. The important reasons for using Dynamic Time Warping are that time-series coming out of the APSOS are not necessarily from the same length and sample rate and these are the characteristics the DTW can handle much better than other distance measures.

DTW tries to find the optimal distance between two time-series by finding the best match between them and to do this the algorithm compares each point of the first time-series to many points of the other. Using this method, time-series with the same patterns occurred in different time periods are considered to be similar. This fact is shown in Figure 3(a) for two time-series with 16 and 13 time-steps. The DTW algorithm for calculating the distance between two given time-series as $TS^{(1)} = \{x_1^{(1)}, x_2^{(1)}, \dots, x_N^{(1)}\}$ and $TS^{(2)} = \{x_1^{(2)}, x_2^{(2)}, \dots, x_M^{(2)}\}$ is presented in Pseudo Code 1**Error! Reference source not found.** [1], [53].

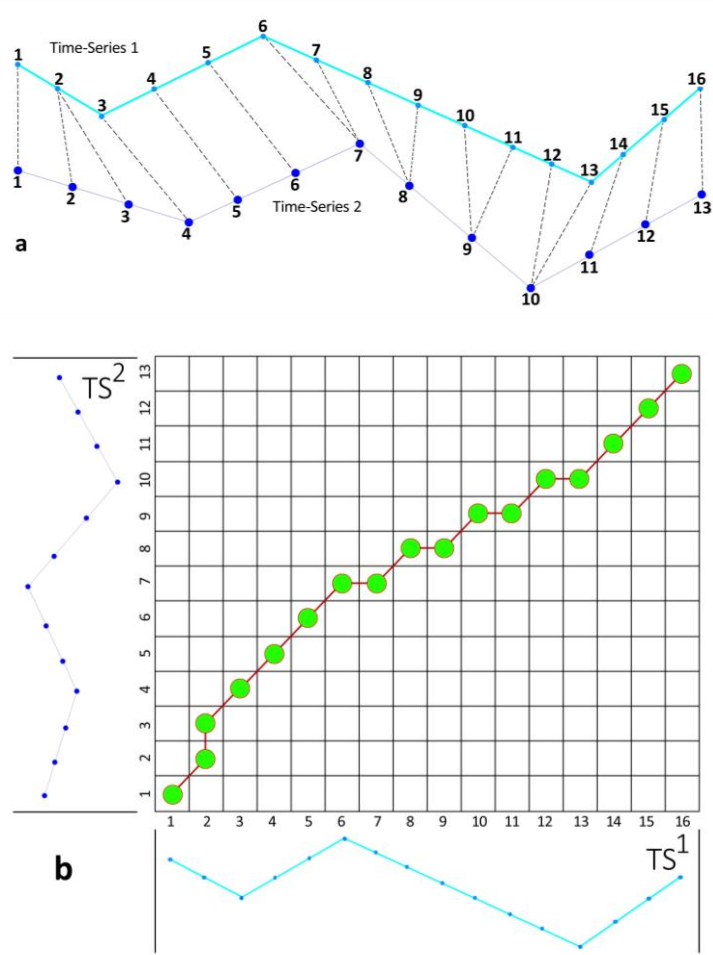

*Figure 3 (a) matching two time-series by DTW (b) the warping path for the time-series matched in (a)*

The function *d(.)* is a distance measure such as Euclidean which is used to calculate the distance between two points each from one sequence. Based on this algorithm, the final distance between the two given series would be $cost_{N,M}$ and the matrix *path* shows the warping path represented in Figure 3(b).

Although DTW is called a distance measure, in fact, it is a method of calculating the optimum distance between two time-series and it is dependent on traditional distance measures such as Euclidean and

**Pseudo Code: Dynamic Time Warping**
$TS^{(1)} = \{x_1^{(1)}, x_2^{(1)}, \ldots, x_N^{(1)}\}$,
$TS^{(2)} = \{x_1^{(2)}, x_2^{(2)}, \ldots, x_M^{(2)}\}$,
$cost_{1,1} = 2d(x_1^{(1)}, x_1^{(2)})$,
$path_{1,1} = (0,0)$,
**for** *i = 2,3,..., N*
   $cost_{i,1} = cost_{i-1,1} + d(x_i^{(1)}, x_1^{(2)})$,
**end for**
**for** *j = 2,3,..., M*
   $cost_{1,j} = cost_{1,j-1} + d(x_1^{(1)}, x_j^{(2)})$,
**end for**
**for** *i = 2,3,..., N*
   **for** *j = 2,3,..., M*
      $cost_{i,j} = \min\{cost_{i,j-1} + d(x_i^{(1)}, x_j^{(2)}), cost_{i-1,j-1} + 2d(x_i^{(1)}, x_j^{(2)}), cost_{i-1,j} + d(x_i^{(1)}, x_j^{(2)})\}$,
      $path_{i,j} = \arg\min_{i,j} cost_{i,j}$,
   **end for**
**end for**

*Pseudo Code 1 Symmetric Dynamic Time Warping algorithm*

Manhattan distances. It is basically admitted in this paper that DTW is the best approach for calculating the distance between two time-series represented by APSOS but the distance measure used inside the DTW must be modified to best adapt with the APSOS outcome. One of the most widely used distance measures which are proved to produce high-quality results for clustering is Euclidean distance from the group of Minkowski distances. The Minkowski distance is defined as

$$d_{Minkowski}\left(x_i^{(1)}, x_j^{(2)}\right) = \left(\left|x_i^{(1)} - x_j^{(2)}\right|^b\right)^{\frac{1}{b}}. \tag{10}$$

Manhattan and Euclidean distances are special cases of Minkowski distance and are obtained by setting $b = 1$ and $b = 2$ respectively [15]. Both of them are widely used within the DTW. But in this paper, a new distance measure based on Euclidean distance and the slope of each segment is developed.

Consider the two time-series of Figure 3(a) are standardized into [-1, 1] and only a part of them is shown in Figure 4(a). Based on Figure 3(a) and Figure 4(a), DTW may match point *12* from the first time-series to point *10* of the second one rather than *9* because $d_1 < d_2$. But in fact, as it is clear from Figure 4 point *12* and *9* are in more similar situations. The reason is that the slope of the segment right after the point *12* is more similar to the one after *9* rather than the one after *10*. The segment after point *10* has a positive slope while the one after *12* has a negative one. This makes points *10* and *12* not from a kind.

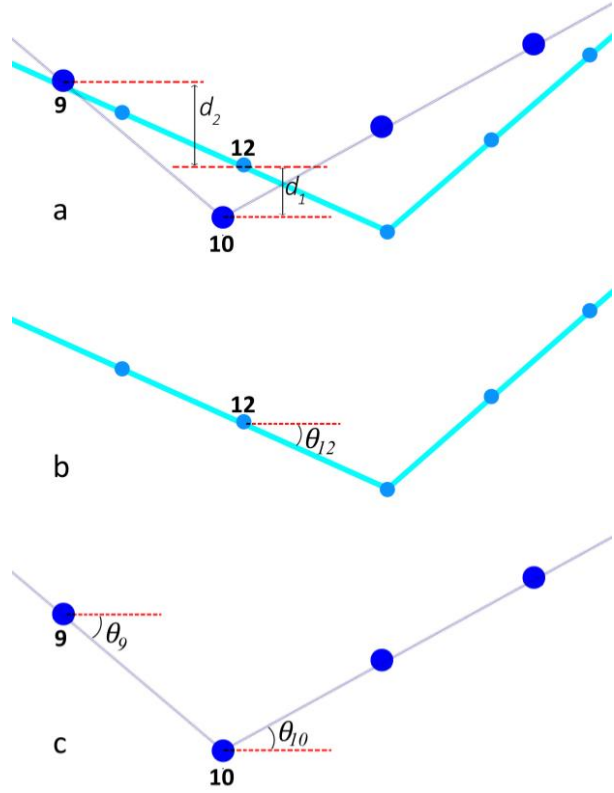

Figure 4 Manhattan distances and angles providing the slope of each segment

This mismatch by DTW is due to its high reliance on the distance measure used inside the algorithm. Manhattan or Euclidean distances are too simple for these cases and they only measure the vertical difference between two points, while in many cases shape characteristics of those points may be very important and effective in matching similar patterns. One of the most important aspects of shape is the slope of a line or a segment which should be implemented inside this distance measure. Following what was said at the beginning of this section about APSOS outcome and how to modify it in order to preserve the slopes, the distance measure used inside the DTW can be a combination of Euclidean distance and slope differences of each segment. For two specific points of two time-series represented in the APSOS form, Figure 5 shows how these values are defined.

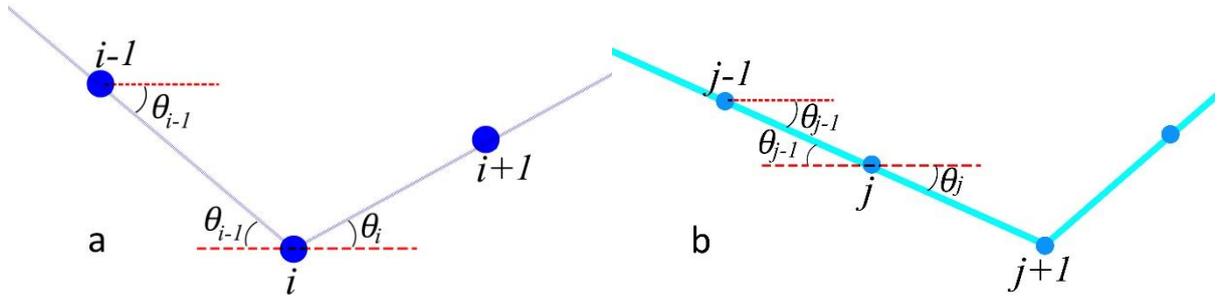

Figure 5 Angle definitions for each point in a time-series

Consider two given time-series as $TS^{(1)} = \{x_1^{(1)}, x_2^{(1)}, \dots, x_N^{(1)}\}$ and $TS^{(2)} = \{x_1^{(2)}, x_2^{(2)}, \dots, x_M^{(2)}\}$. The slope-based distance between two points of these time-series can be calculated by

$$d_{slope-based}\left(TS_i^{(1)}, TS_j^{(2)}\right) = d_{Euclidean} + \left|\sin\theta_i^{(1)} - \sin\theta_j^{(2)}\right|, \tag{11}$$

where $\theta_i$ are defined as in Figure 5 and $d_{Euclidean}$ is the Euclidean distance between the $i^{th}$ point of the first time-series and the $j^{th}$ point of the second one, calculated by Eq. (10) when $b = 2$. Hence the simple slop-based distance between two points will be

$$d_{slope-based}\left(TS_i^{(1)}, TS_j^{(2)}\right) = \sqrt{\left(x_i^{(1)} - x_j^{(2)}\right)^2 + \left|\sin\theta_i^{(1)} - \sin\theta_j^{(2)}\right|}. \tag{12}$$

Simply it could be understood that considering only the slope of the segment on the right side of point will not be sufficient. It happens especially when one of the points is a local extremum and the other one is not. This time, the slopes of segments on both sides should be taken into consideration. Thus, the bilateral slope-based distance (BSD) will be

$$d_{BSD}\left(TS_i^{(1)}, TS_j^{(2)}\right) = \left|x_i^{(1)} - x_j^{(2)}\right| + \left|\sin\theta_i^{(1)} - \sin\theta_j^{(2)}\right| + \left|\sin\theta_{i-1}^{(1)} - \sin\theta_{j-1}^{(2)}\right|. \tag{13}$$

It must be mentioned that to standardize the distance measure, before using it, both time-series should be standardized into [-1, 1]. Thus $-1 \leq x_i^{(1)}, x_i^{(2)} \leq 1$ and therefore $0 \leq \left|x_i^{(1)} - x_j^{(2)}\right| \leq 2$. Since $-1 \leq \sin\theta_i^{(1)}, \sin\theta_j^{(2)} \leq 1$ thus $0 \leq \left|\sin\theta_i^{(1)} - \sin\theta_j^{(2)}\right| \leq 2$. This means that each term of the BSD distance measure is in the same interval of [0, 2], which is also the reason behind selecting the sine as the representative of the slope, making it independent of any weighting coefficients. For the first and the last points of the time-series, only one term will be calculated and therefore the simple form of BSD is used. Based on Eq. (13) the BSD function is defined on $\mathbb{R}^3$ so $d_{BSD}: \mathbb{R}^3 \times \mathbb{R}^3 \to \mathbb{R}$.

Since using a metric distance measure is desired as it is easier to index the space for speed-up search, it is necessary to show if the proposed distance measure is a metric one.

**Definition 3:** a pair of $(S, d)$ is called metric space iff

i) $S$ is a set,
ii) $d: S \times S \to \mathbb{R}$ is a function with following four properties for all $x, y, z$ in $S$:
   (P1)     $d(x, y) \geq 0$     (nonnegativity)
   (P2)     $d(x, y) = 0 \Leftrightarrow x = y$     (identity)
   (P3)     $d(x, y) = d(y, x)$     (symmetry)
   (P4)     $d(x, y) \leq d(x, z) + d(y, z)$     (triangle inequality).

Based on the definition of a metric space, the distance function should satisfy the four aforementioned properties in order to be called a metric distance measure by definition [59][38][60].

**Proposition 1:** the pair $(\mathbb{R}^3, d_{BSD})$ is metric space ($\equiv$ Bilateral Slope-based Distance is a metric distance function)

**Proof:** see the Appendix.

Since the proposed distance measure is metric, indexing can be used which is very crucial for having efficiency on data mining tasks such as clustering especially for large databases (it should be mentioned that Dynamic Time Warping does not necessarily obey the triangle inequality (P4), even if its local distance measure is metric.)

In the next section, PSO clustering will be briefly explained and then in section 5, the proposed distance measure is compared to some well-known distance measures.

## 4. Particle Swarm Optimization Clustering

PSO, an evolutionary and population-based algorithm in the field of swarm intelligence, has been developed as a heuristic technique for dealing with optimization problems [21], [22], [61], [62]. This technique is inspired by the natural and biological behavior of a group of animals like bees, ants, fish and birds which are collectively searching for food and communicating with each other [20], [24], [61]. On the basis of this natural behavior, in this technique, a swarm of particles, which is a representative of the population of animals, moves in the solution space, while each individual -each particle- tries to find an optimal solution and then cooperates and communicates with others by sharing its best solution [15].

To better understand the technique, consider a predefined objective function, $F$, which is desired to be optimized by minimization. Supposing the solution to this function is $n$-dimensional, the solution space would be $n$-dimensional [21]. In order to find the optimal solution for this function, a swarm of $P$ particles is released into the solution space while each particle $p$ is a candidate for the solution and is distinguished from other particles by two vectors called position vector, $X_p^{(iter)}$ and velocity vector, $V_p^{(iter)}$ which are all from the same dimension of the solution space. The position vector is a solution and the velocity vector is used to move the particle to its next position which is going to be the next solution. In each iteration *(iter)*, particles' positions are fed to the objective function and they are compared to each other's best positions ever and the best position of all the particles until that iteration. These two are called respectively personal best position, $X_p^{PB(iter)}$ and global best position, $X^{GB(iter)}$. For the next iteration, the velocity of each particle must be updated in a way that the particle not only moves toward the global best position but also toward its personal best position. Then the new position of each particle is updated and again these positions which are in fact new solutions for the objective function, are fed into the function and they are compared. Through a number of iterations, these particles evolve and find an optimal solution.

In order to match PSO to the clustering of time series, some modifications need to be applied on PSO. First of all, for the similarity or distance measure denoted by $d(.)$ in this paper, a distance measure which is compatible with time-series should be employed. For the objective function used both in PSO algorithm and clustering, the combined measure is selected to involve both the compactness and separation measures. Here, the objective function in PSO is the fitness function in clustering. Therefore, the goal of the PSO algorithm here is to minimize the fitness function so that the quality of clusters increases. To do so, a swarm of $P$ particles with random values for Particle's positions and velocity vectors is released into the search space. Each particle's position is a clustering solution containing the cluster centers. Therefore, the position vector of $p^{th}$ particle at iteration *iter* represents $K$ cluster centers as below

$$X_p^{(iter)} = (m^1, \ldots, m^K)^{(iter)}, p = 1, \ldots, P \qquad (14)$$

where $m^k$ is the center of cluster $k$ or cluster prototype, which here is a time series defined by the medoid sequence approach described in section 2. During each iteration, first the velocity vector and then the position vector of each particle is updated respectively according to Eq. (15) and Eq. (16):

$$V_p^{(iter)} = wV_p^{(iter-1)} + c_1 r_1 \left(X_p^{PB(titer-1)} - X_p^{(iter-1)}\right) + c_2 r_2 \left(X_p^{GB(iter-1)} - X_p^{(titer-1)}\right), \qquad (15)$$

$$X_p^{(iter)} = X_p^{(iter-1)} + V_p^{(iter)}. \qquad (16)$$

The $w$ in the velocity update statement is called the inertia weight and it shows the impact of the previous velocity of a particle on the current one. Also, $c_1$ and $c_2$ are respectively cognitive coefficient and social coefficient; the first one indicates the tendency of a particle to move toward its best previous positions and the second one indicates the tendency of a particle to follow the best previous position of the whole swarm. The $r_1$ and $r_2$ coefficients are numbers from the uniform distribution in interval [0, 1] selected randomly [15], [20], [21].

Then the time series in data set are one by one assigned to the nearest cluster center which is defined in the new position vector and this is done for each particle. The reason is that each particle is a distinct solution and therefore this assigning must be done for each particle. Then the fitness function for each particle is calculated and these values are compared and then the values of personal best position and global best position are updated.

The termination condition can be a maximum number of iterations, number of iterations without improvement and a minimum objective function criterion [21]. This paper uses 500 as the maximum number of iterations for the termination condition.

**Pseudo Code: PSO-based Algorithm for Clustering Time-Series Data**
Initialize a swarm of P particles,
*//initial positions and velocities*
**for** each particle *p*
  $X_p^{(0)} = a\ random\ value$ for each $m^k$
  $V_p^{(0)} = a\ random\ value,$
  $X_p^{PB(0)} = X_p^{(0)},$
  *//initial assigning of time series to clusters*
  **for** each time series *TSj* in data set
    **for** each cluster center *k*
      calculate $d(TSj, m^k),$
    **end for**
    assign *TSj* to the nearest cluster
  **end for**
  $X^{GB(0)} = arg \min_{X_p^{PB(0)}} F_{Combined}(X_p^{PB(0)}),$
**end for**
iter = 1,
**while** (the termination condition is not met)
  **for** *p* = 1 to P
    *//velocity and position update:*
    $V_p^{(iter)} = wV_p^{(iter-1)} + c_1 r_1 \left( X_p^{PB(iter-1)} - X_p^{(iter-1)} \right) + c_2 r_2 \left( X_p^{GB(iter-1)} - X_p^{(iter-1)} \right),$
    $X_p^{(iter)} = X_p^{(iter-1)} + V_p^{(iter)},$
    *//assigning of time series to new clusters*
    **for** each time series *TSj* in data set
      **for** each cluster center *k*
        calculate $d(TSj, m^k),$
      **end for**
      assign *TSj* to the nearest cluster
    **end for**
    *//personal best position update:*
    **if** $F_{Combined}(X_p^{(iter)}) < F_{Combined}(X_p^{PB(iter-1)})$
      $X_p^{PB(iter)} = X_p^{(iter)},$
      *//global best position update:*
      **if** $F_{Combined}(X_p^{PB(iter)}) < F_{Combined}(X^{GB(iter-1)})$
        $X^{GB(iter-1)} = X_p^{PB(iter)},$
      **end if**
    **end if**
  **end for**
**end while**
iter = iter+1, *//next iteration*

*Pseudo Code 2 PSO-based algorithm for clustering time-series data*

Through a number of iterations and when the termination condition is met, the optimal solution which is the last global best position is reached. The algorithm of the proposed method for clustering time series data in this paper is described in Pseudo Code 2.

The choice of PSO algorithm to perform the task of clustering is highly affected by the fact that this algorithm and more generally the evolutionary algorithms do a very good job in complex and large-scale optimization problems. Many studies have shown that among the evolutionary algorithms, PSO is the one that outperforms others in most of the metrics considered to be important in the field of optimization [63,64]. PSO outperforms most of the evolutionary algorithms in terms of the success rate, and the processing time [63]. It should be mentioned that the choice of PSO in this paper as the algorithm that performs the task of clustering is not a contribution of this paper since many other evolutionary algorithms can be used instead of PSO to do so. In fact, the modification of the algorithm in a way that complies with the problem of time-series clustering and can work with various similarity measures requires a huge effort and not always is supposed to work. This paper then uses a standard single-swarm PSO which works based on social learning.

## 5. Experimental Results and Discussion

This section of the paper provides the experimental results of testing and comparing the proposed distance measure to other methods of their kind on some time-series data sets. The data sets used in this paper are from the UCR time-series classification archive [65] and are available for free. Nine data sets are selected from this archive and their characteristics are provided in Table 1. The reason for selecting them is that they have the longest lengths and more samples among the rest of the datasets. The test sets have been used for the experiments. Figure 6 shows one sample of each class for these data sets.

*Table 1 Selected data sets from the UCR time-series archive*

| Dataset Name | $T$ of data set | Time-series length($N$) | No. of Classes |
|---|---|---|---|
| CinC_ECG_torso | 1420 | 1639 | 4 |
| HandOutlines | 1370 | 2709 | 2 |
| Haptics | 463 | 1092 | 5 |
| InlineSkate | 650 | 1882 | 7 |
| MALLAT | 2400 | 1024 | 8 |
| Phoneme | 2110 | 1024 | 39 |
| StarLightCurves | 9236 | 1024 | 3 |
| UWaveGestureLibrary | 4478 | 945 | 8 |
| Worms | 258 | 900 | 5 |

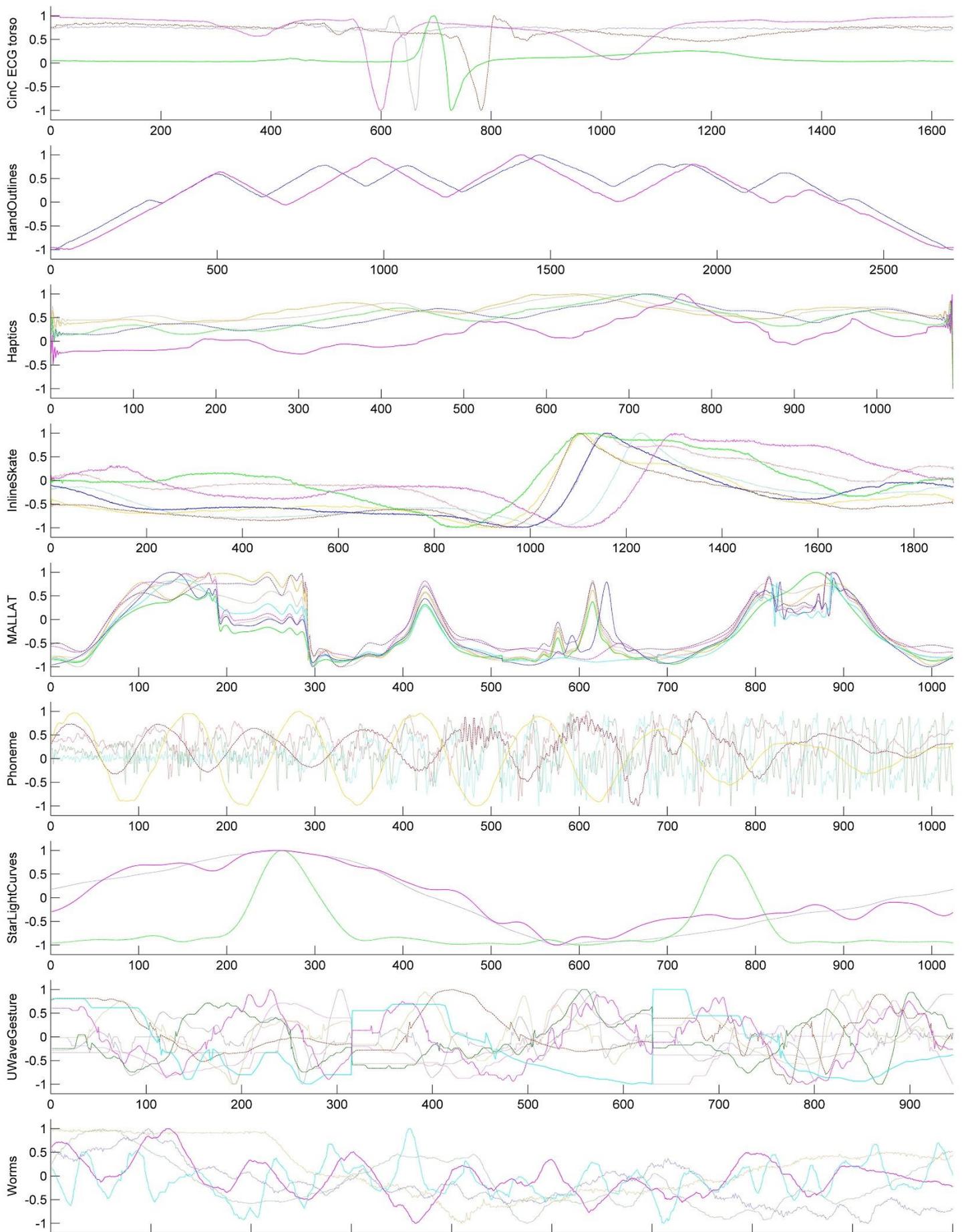

*Figure 6 Samples from each class of the 9 data sets*

These data sets were given to the APSOS algorithm in order to not only reduce their dimensions but also to make them have a better representation for the experiments. The changes in the average of time-series' lengths after applying APSOS is represented in Table 2. These time-series were all standardized into [-1, 1].

*Table 2 Changes in the time-series' lengths by applying APSOS*

| Dataset Name | Raw Time-series length | Avg. Reduced Length |
|---|---|---|
| CinC_ECG_torso | 1639 | 154 |
| HandOutlines | 2709 | 178 |
| Haptics | 1092 | 42 |
| InlineSkate | 1882 | 139 |
| MALLAT | 1024 | 127 |
| Phoneme | 1024 | 36 |
| StarLightCurves | 1024 | 155 |
| UWaveGestureLibrary | 945 | 82 |
| Worms | 900 | 231 |

Since the performance of PSO-based clustering algorithm has been proved in other researches [15], [21], [22], [66], in this section of the paper only the proposed distance measure is compared to 3 other distance measures widely used for time-series similarity searches, namely Dynamic Time Warping (DTW) with Euclidean Distance [53], Edit Distance on Real Sequence (EDR) [44], and Longest Common Subsequence (LCSS) [67]. It should be mentioned that the proposed distance measure in this paper, BSD, is used with DTW to be compatible with the varied length time-series produced by any dimensionality reduction technique. The PSO-based algorithm presented in Pseudo Code 2 is used to conduct the task of clustering for all the mentioned distance measures. For the cluster prototypes, the medoid sequence approach is taken. The combined function described in section 2 is selected as the fitness function as it considers both the separation and compactness of the clusters. Figure 7 shows the performance of the mentioned distance measures on each data set in terms of the combined criteria.

It is important to notice that all results are obtained by repeating the algorithms 10 times and for each experiment, the PSO parameters are set to $w = 1.2$, $c_1 = 1.5$, $c_2 = 1.5$, and $n=30$ (these parameters have been tuned using Taguchi method[68]). The $w$ should decrease gradually through iterations.

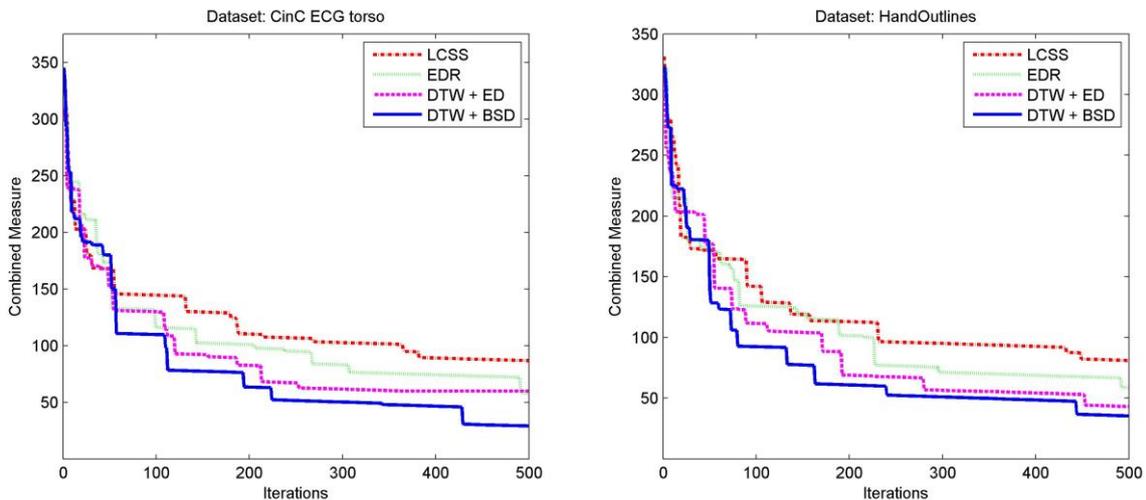

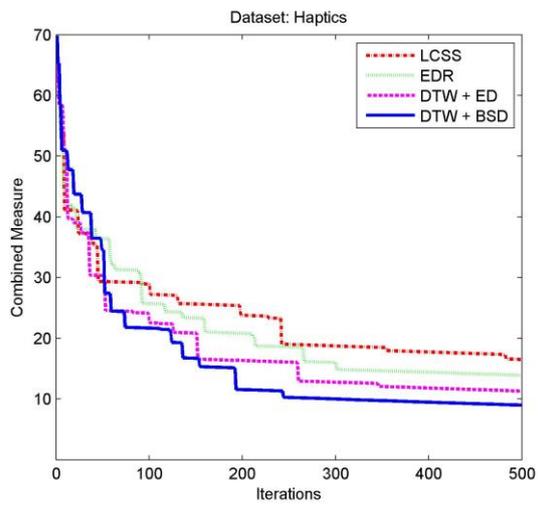
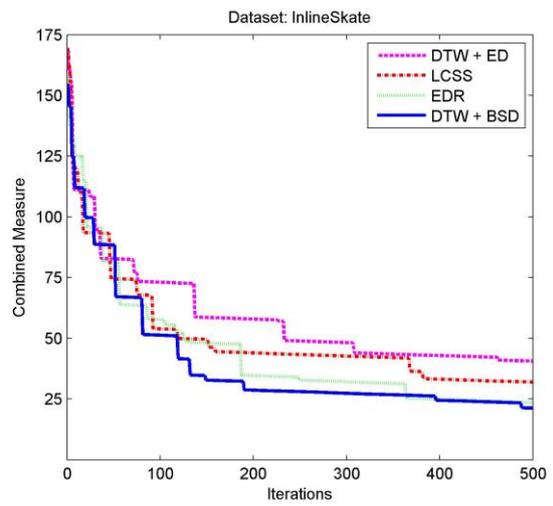
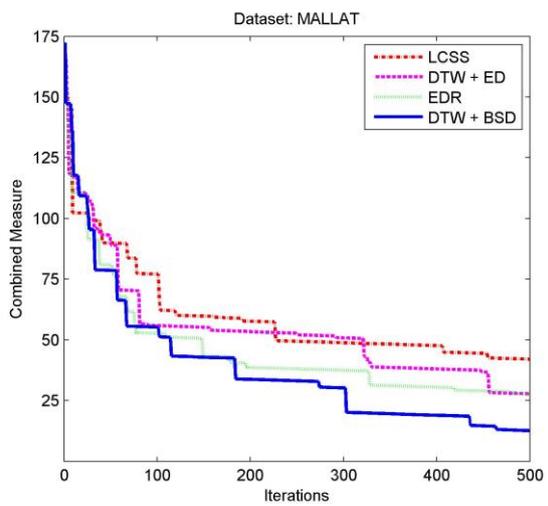
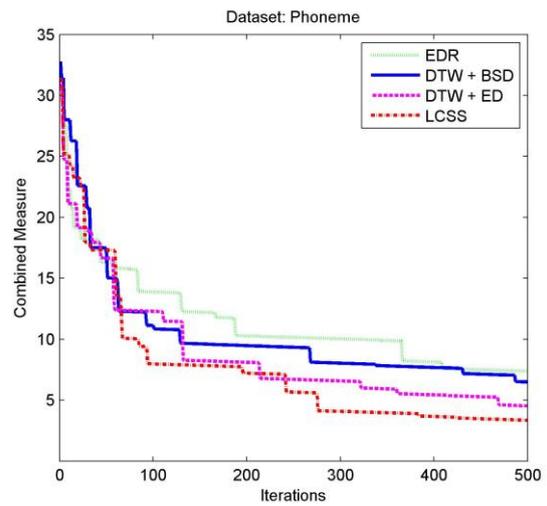
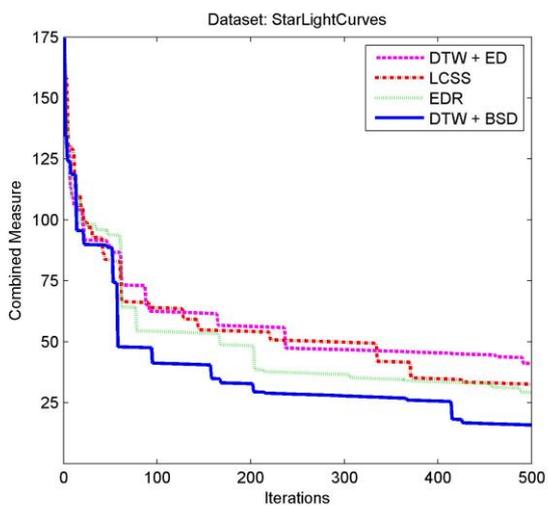
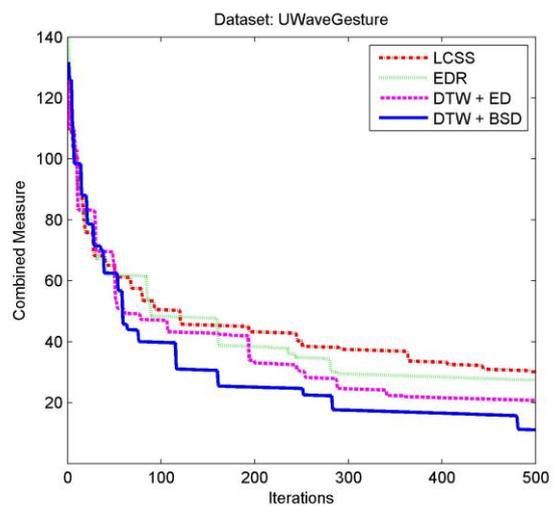

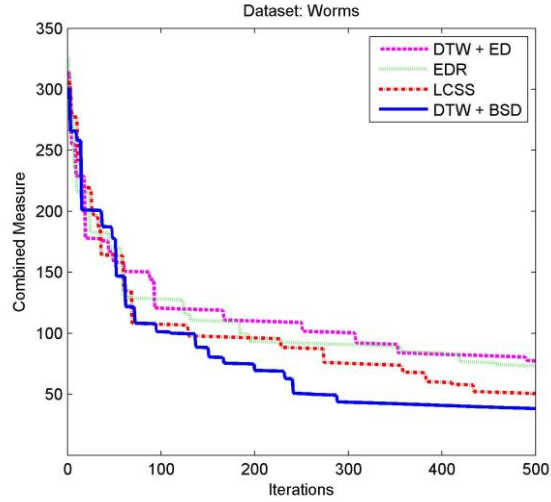

*Figure 7 Results for clustering with each distance measure in terms of the combined measure*

As it is clear from the results for 9 datasets in Figure 7, the BSD measure combined with DTW outperforms the rest of the measures in 8 datasets and improves the quality of clustering more than the rest of the measures. It must be mentioned that the selected measures are among the best and their performances are usually very close to each other. The reason behind the BSD's lack of good performance in dataset *Phoneme* may rely on the fact that dimensionality reduction technique is weak in transforming a high-frequency time-series and most of the time-series in the dataset *Phoneme* are from high-frequency.

The final results of clustering are also evaluated by other criteria mentioned in section 2, such as Cluster Purity, Rand Index, and Jaccard Score. They are shown in Table 3 to Table 11. The bold ones are the best in each criterion (the letter σ indicates the standard deviation).

*Table 3 Average and standard deviation of different measures for CinC ECG torso dataset*

| Distance Measure | Cluster Purity [σ] | Jaccard Score [σ] | Rand Index [σ] |
|---|---|---|---|
| LCSS | 0.7562 [0.0553] | 0.3055 [0.1396] | 0.6382 [0.1672] |
| EDR | 0.8011 [0.0632] | 0.3292 [0.0809] | 0.6646 [0.1229] |
| DTW + ED | 0.8629 [0.0879] | 0.3753 [0.1343] | 0.7008 [0.1819] |
| DTW + BSD | **0.9303** [0.0418] | **0.4248** [0.1250] | **0.7240** [0.1432] |

*Table 4 Average and standard deviation of different measures for HandOutlines dataset*

| Distance Measure | Cluster Purity [σ] | Jaccard Score [σ] | Rand Index [σ] |
|---|---|---|---|
| LCSS | 0.7613 [0.0958] | 0.3313 [0.0985] | 0.6646 [0.1088] |
| EDR | 0.8448 [0.0795] | 0.3820 [0.0871] | **0.8892** [0.0966] |
| DTW + ED | 0.8717 [0.0822] | 0.4497 [0.1394] | 0.7863 [0.0777] |
| DTW + BSD | **0.9231** [0.0556] | **0.4553** [0.0952] | 0.8604 [0.1063] |

*Table 5 Average and standard deviation of different measures for Haptics dataset*

| Distance Measure | Cluster Purity [σ] | Jaccard Score [σ] | Rand Index [σ] |
|---|---|---|---|
| LCSS | 0.6999 [0.0880] | 0.3188 [0.1190] | 0.6500 [0.1373] |
| EDR | 0.7134 [0.0667] | 0.3216 [0.1210] | 0.6266 [0.1457] |
| DTW + ED | 0.7491 [0.0183] | 0.3367 [0.1001] | 0.6634 [0.1586] |
| DTW + BSD | **0.8399** [0.0128] | **0.3814** [0.1262] | **0.7458** [0.1788] |

*Table 6 Average and standard deviation of different measures for InlineSkate dataset*

| Distance Measure | Cluster Purity [σ] | Jaccard Score [σ] | Rand Index [σ] |
|---|---|---|---|
| LCSS | 0.7922 [0.0624] | 0.3967 [0.1203] | 0.7036 [0.1309] |
| EDR | 0.8147 [0.0342] | 0.4661 [0.1416] | 0.6526 [0.1191] |
| DTW + ED | 0.7530 [0.0878] | 0.4458 [0.1204] | 0.6257 [0.1418] |
| DTW + BSD | **0.8291** [0.0722] | **0.4815** [0.1052] | **0.7157** [0.1833] |

*Table 7 Average and standard deviation of different measures for MALLAT dataset*

| Distance Measure | Cluster Purity [σ] | Jaccard Score [σ] | Rand Index [σ] |
|---|---|---|---|
| LCSS | 0.7288 [0.0796] | 0.3488 [0.0723] | 0.6770 [0.1456] |
| EDR | 0.8132 [0.0988] | 0.4188 [0.1096] | 0.6121 [0.1660] |
| DTW + ED | 0.7801 [0.1069] | **0.4716** [0.1006] | 0.6016 [0.1189] |
| DTW + BSD | **0.8791** [0.0770] | 0.4568 [0.0573] | **0.6794** [0.1645] |

*Table 8 Average and standard deviation of different measures for Phoneme dataset*

| Distance Measure | Cluster Purity [σ] | Jaccard Score [σ] | Rand Index [σ] |
|---|---|---|---|
| LCSS | **0.7880** [0.1064] | 0.3501 [0.1461] | **0.7413** [0.1385] |
| EDR | 0.6833 [0.1130] | **0.3576** [0.1267] | 0.7161 [0.1532] |
| DTW + ED | 0.7745 [0.0699] | 0.3211 [0.1270] | 0.7002 [0.1412] |
| DTW + BSD | 0.7194 [0.0828] | 0.3037 [0.1197] | 0.6523 [0.1203] |

*Table 9 Average and standard deviation of different measures for StarLightCurves dataset*

| Distance Measure | Cluster Purity [σ] | Jaccard Score [σ] | Rand Index [σ] |
|---|---|---|---|
| LCSS | 0.8266 [0.0665] | 0.3758 [0.1433] | 0.6048 [0.1539] |
| EDR | 0.8327 [0.0881] | 0.3105 [0.1112] | 0.6569 [0.1299] |
| DTW + ED | 0.8037 [0.0873] | 0.3974 [0.1042] | 0.7130 [0.1722] |
| DTW + BSD | **0.8919** [0.0427] | **0.4508** [0.1146] | **0.7958** [0.1861] |

*Table 10 Average and standard deviation of different measures for UWaveGesture dataset*

| Distance Measure | Cluster Purity [σ] | Jaccard Score [σ] | Rand Index [σ] |
|---|---|---|---|
| LCSS | 0.7589 [0.1018] | 0.4213 [0.1024] | 0.7434 [0.1535] |
| EDR | 0.7864 [0.0838] | 0.3615 [0.1333] | **0.7472** [0.1610] |
| DTW + ED | 0.8414 [0.0584] | 0.4221 [0.1307] | 0.6605 [0.1827] |
| DTW + BSD | **0.8843** [0.0263] | **0.4570** [0.1521] | 0.6626 [0.1261] |

*Table 11 Average and standard deviation of different measures for Worms dataset*

| Distance Measure | Cluster Purity [σ] | Jaccard Score [σ] | Rand Index [σ] |
|---|---|---|---|
| LCSS | 0.8716 [0.0672] | 0.3043 [0.1004] | 0.8209 [0.1290] |
| EDR | 0.7883 [0.0619] | **0.3853** [0.0758] | 0.7711 [0.1528] |
| DTW + ED | 0.7253 [0.1163] | 0.3183 [0.1433] | 0.7298 [0.1186] |
| DTW + BSD | **0.9146** [0.0258] | 0.3468 [0.1137] | **0.8574** [0.1431] |

Also, the final results are evaluated using the SSE measure and represented in Figure 8.

Based on the results acquired by various measures mentioned and shown above, DTW which was shown to outperform distances such as EDR and LCSS works better with BSD rather than the simple Euclidean Distance.

As the choice of the similarity measure can affect the time it takes algorithms to perform the clustering,

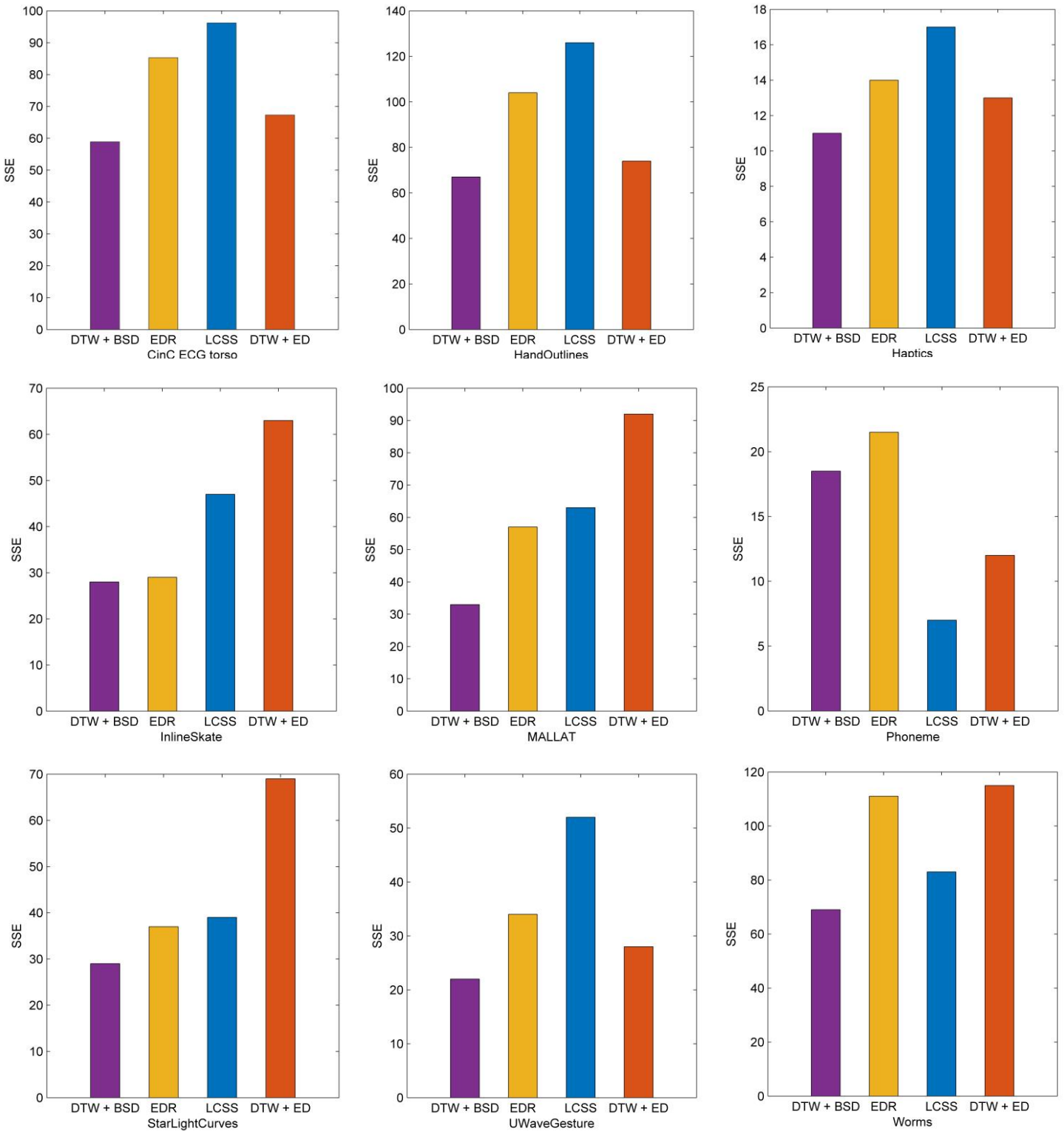

*Figure 8 Average SSE of different measures for each dataset*

a time analysis has also been conducted to show the comparison between the mentioned similarity measures. Table 12 demonstrates the time it takes for PSO to reach 500 iterations using each of the similarity measures in each dataset. As the Table 12 shows, in 4 out of 9 datasets, PSO with the proposed distance measure, performs the clustering faster than other similarity measures and in the second place stands DTW + ED with being the fastest in 3 out of 9 datasets. Though it seems that PSO with the proposed distance measure is doing a better job in terms of time than other similarity measures, the times are so close that a strict conclusion cannot be made based on the results in Table 12.

*Table 12 The time it takes for PSO to reach 500 iterations using each similarity measure in each dataset in minutes [σ]*

| Distance Measure | CinC_ECG_torso | HandOutlines | Haptics | InlineSkate | MALLAT | Phoneme | StarLightCurves | UWaveGestureLibrary | Worms |
|---|---|---|---|---|---|---|---|---|---|
| LCSS | 2.15 [0.031] | 3.69 [0.073] | 1.78 [0.093] | 2.60 [0.027] | 4.21 [0.043] | 2.83 [0.012] | 11.23 [0.044] | **5.38** [0.077] | 3.37 [0.045] |
| EDR | 2.11 [0.024] | **3.21** [0.065] | 1.76 [0.048] | 2.20 [0.033] | 4.47 [0.084] | 3.14 [0.003] | 11.67 [0.029] | 6.26 [0.065] | 3.76 [0.05] |
| DTW + ED | 2.40 [0.027] | 3.45 [0.027] | 1.79 [0.009] | **2.19** [0.021] | 4.41 [0.035] | **2.57** [0.02] | 10.89 [0.075] | 5.50 [0.046] | **3.03** [0.037] |
| DTW + BSD | **1.95** [0.041] | 3.56 [0.033] | **1.75** [0.092] | 2.45 [0.026] | **4.11** [0.092] | 2.95 [0.054] | **10.81** [0.048] | 5.99 [0.034] | 3.12 [0.064] |

# 6. Conclusion

In this paper first the deficiency of most of the representation methods in preserving the original time-series' shapes was removed by modifying the output of the APSOS algorithm and letting it preserve the slope of each segment in the form of the sine of the angle each segment makes with the horizontal line passing from the starting point of the segment. Since this representation method is a shape-based one and is designed to preserve the shape-characteristics of a given time-series and also to reduce the dimension of it, a new distance measure called Bilateral Slope-based Distance (BSD) was developed based on the output of this representation method, and the two so-called distance measures, Euclidean distance and DTW. The developed distance measure is based on the fact that when two points are similar in shape, the slope of the segments adjacent to the first point should be similar to the respective slopes adjacent to the second point. This similarity is based on the shape of the breakpoints or extremums of a time-series. Since this distance measure is based on the representation method used and usually representation methods do not have the same sample rate from the original time-series, the proposed distance measure should be used with one of the elastic measures such as dynamic time warping. Thus, in this paper, the proposed distance measure is combined with DTW and it is used instead of the simple Euclidean distance within the DTW. This way DTW not only considers the vertical difference of the time-series points but also the real shape differences of each pair. BSD can be called a shape-based distance measure for time-series and while used with DTW can best calculate the difference (or similarity) between two time-series. The results of the experiments conducted on various data sets with four different distance measures including DTW with BSD, DTW with ED, EDR, and LCSS show that BSD used along with DTW improves the clustering results more than the other mentioned distance measures, not only in one criterion but in five different measures. Further studies could focus on how to define weighting coefficients for the proposed distance measure to discriminate between each slope and to find the importance of each one. The BSD still uses the Euclidean distance between two points as a component of it. Further studies could also investigate the use of other distances instead of the Euclidean one to see the effect on the result of clustering. The suggestion is to use those distances that capture the shape of time-series the most. BSD captures the slope of each segment as the sine of the angles it makes. Other forms of capturing these slopes can also be investigated in further studies.

# Appendix
**Proof for Proposition 1:** assume any $x, y, z \in \mathbb{R}^3$ such that

$$x = TS_i^{(1)} = (x_i^{(1)}, \theta_i^{(1)}, \theta_{i-1}^{(1)}),$$

$$y = TS_j^{(2)} = (x_j^{(2)}, \theta_j^{(2)}, \theta_{j-1}^{(2)}),$$

$$z = TS_k^{(3)} = (x_k^{(3)}, \theta_k^{(3)}, \theta_{k-1}^{(3)}),$$

where, as defined before, e.g. $TS_i^{(1)}$ is a representative of the $i^{th}$ point in Time-series (1) depicted by a triplet of the point's value ($x_i^{(1)}$) and its respective angles ($\theta_i^{(1)}$ and $\theta_{i-1}^{(1)}$).

It is clear that $\mathbb{R}^3$ is a set. So part (i) of the definition 3 is satisfied.

For part (ii) of the definition 3, function $d_{BSD}: \mathbb{R}^3 \times \mathbb{R}^3 \to \mathbb{R}$ should satisfy properties 1 to 4.

For property 1:

$$d_{BSD}(x,y) = d_{BSD}\left(TS_i^{(1)}, TS_j^{(2)}\right) = \underbrace{\left|x_i^{(1)} - x_j^{(2)}\right|}_{\geq 0} + \underbrace{\left|\sin\theta_i^{(1)} - \sin\theta_j^{(2)}\right|}_{\geq 0} + \underbrace{\left|\sin\theta_{i-1}^{(1)} - \sin\theta_{j-1}^{(2)}\right|}_{\geq 0} \geq 0.$$

Since all the terms are nonnegative, $d_{BSD}(x,y) \geq 0$.

For property 2: assuming $d_{BSD}(x,y) = 0$, then

$$d_{BSD}(x,y) = d_{BSD}\left(TS_i^{(1)}, TS_j^{(2)}\right) = \underbrace{\left|x_i^{(1)} - x_j^{(2)}\right|}_{=0} + \underbrace{\left|\sin\theta_i^{(1)} - \sin\theta_j^{(2)}\right|}_{=0} + \underbrace{\left|\sin\theta_{i-1}^{(1)} - \sin\theta_{j-1}^{(2)}\right|}_{=0} = 0.$$

Since all terms are nonnegative, the only way that the above equation can be zero is when all terms are zero, thus

$$\left|x_i^{(1)} - x_j^{(2)}\right| = 0 \implies x_i^{(1)} = x_j^{(2)},$$

$$\left|\sin\theta_i^{(1)} - \sin\theta_j^{(2)}\right| = 0 \implies \sin\theta_i^{(1)} = \sin\theta_j^{(2)} \xrightarrow{-\pi/2 < \theta_i^{(1)}, \theta_j^{(2)} < \pi/2} \theta_i^{(1)} = \theta_j^{(2)},$$

The same applies to $\theta_{i-1}^{(1)}$ and $\theta_{j-1}^{(2)}$. Thus $\left(x_i^{(1)}, \theta_i^{(1)}, \theta_{i-1}^{(1)}\right) = \left(x_j^{(2)}, \theta_j^{(2)}, \theta_{j-1}^{(2)}\right)$ and therefore $TS_i^{(1)} = TS_j^{(2)}$ which means $x = y$.

Now for the reverse in property 2, assume $x = y$, thus

$TS_i^{(1)} = TS_j^{(2)}$ and then $\left(x_i^{(1)}, \theta_i^{(1)}, \theta_{i-1}^{(1)}\right) = \left(x_j^{(2)}, \theta_j^{(2)}, \theta_{j-1}^{(2)}\right)$. Thus

$$d_{BSD}(x,y) = d_{BSD}\left(TS_i^{(1)}, TS_j^{(2)}\right) = \left|x_i^{(1)} - x_j^{(2)}\right| + \left|\sin\theta_i^{(1)} - \sin\theta_j^{(2)}\right| + \left|\sin\theta_{i-1}^{(1)} - \sin\theta_{j-1}^{(2)}\right| =$$
$$\underbrace{\left|x_i^{(1)} - x_i^{(1)}\right|}_{=0} + \underbrace{\left|\sin\theta_i^{(1)} - \sin\theta_j^{(2)}\right|}_{=0} + \underbrace{\left|\sin\theta_{i-1}^{(1)} - \sin\theta_{j-1}^{(2)}\right|}_{=0} = 0.$$

For property 3:

$$d_{BSD}(x,y) = d_{BSD}\left(TS_i^{(1)}, TS_j^{(2)}\right) = \left|x_i^{(1)} - x_j^{(2)}\right| + \left|\sin\theta_i^{(1)} - \sin\theta_j^{(2)}\right| + \left|\sin\theta_{i-1}^{(1)} - \sin\theta_{j-1}^{(2)}\right| =$$
$$\left|x_j^{(2)} - x_i^{(1)}\right| + \left|\sin\theta_i^{(1)} - \sin\theta_j^{(2)}\right| + \left|\sin\theta_{i-1}^{(1)} - \sin\theta_{j-1}^{(2)}\right| = d_{BSD}\left(TS_j^{(2)}, TS_i^{(1)}\right) = d_{BSD}(y, x).$$

For property 4: based on the triangle inequality that says $|v + u| \leq |v| + |u|$, it is clear that adding and subtracting elements from $z$ in each corresponding term of $d_{BSD}(x, y)$ will result in:

$$\left|\underbrace{x_i^{(1)} - x_k^{(3)}}_{} + \underbrace{x_k^{(3)} - x_j^{(2)}}_{}\right| \leq \left|x_i^{(1)} - x_k^{(3)}\right| + \left|x_k^{(3)} - x_j^{(2)}\right|,$$

$$\left|\underbrace{\sin\theta_i^{(1)}-\sin\theta_k^{(3)}}+\underbrace{\sin\theta_k^{(3)}-\sin\theta_j^{(2)}}\right|\le\left|\sin\theta_i^{(1)}-\sin\theta_k^{(3)}\right|+\left|\sin\theta_k^{(3)}-\sin\theta_j^{(2)}\right|,$$

$$\left|\underbrace{\sin\theta_{i-1}^{(1)}-\sin\theta_{k-1}^{(3)}}+\underbrace{\sin\theta_{k-1}^{(3)}-\sin\theta_{j-1}^{(2)}}\right|\le\left|\sin\theta_{i-1}^{(1)}-\sin\theta_{k-1}^{(3)}\right|+\left|\sin\theta_{k-1}^{(3)}-\sin\theta_{j-1}^{(2)}\right|,$$

thus,

$d_{BSD}(x,y)=d_{BSD}\left(TS_i^{(1)},TS_j^{(2)}\right)=\left|x_i^{(1)}-x_j^{(2)}\right|+\left|\sin\theta_i^{(1)}-\sin\theta_j^{(2)}\right|+\left|\sin\theta_{i-1}^{(1)}-\sin\theta_{j-1}^{(2)}\right|=\left|x_i^{(1)}-x_k^{(3)}+x_k^{(3)}-x_j^{(2)}\right|+\left|\sin\theta_i^{(1)}-\sin\theta_k^{(3)}+\sin\theta_k^{(3)}-\sin\theta_j^{(2)}\right|+\left|\sin\theta_{i-1}^{(1)}-\sin\theta_{k-1}^{(3)}+\sin\theta_{k-1}^{(3)}-\sin\theta_{j-1}^{(2)}\right|\le\left|x_i^{(1)}-x_k^{(3)}\right|+\left|x_k^{(3)}-x_j^{(2)}\right|+\left|\sin\theta_i^{(1)}-\sin\theta_k^{(3)}\right|+\left|\sin\theta_k^{(3)}-\sin\theta_j^{(2)}\right|+\left|\sin\theta_{i-1}^{(1)}-\sin\theta_{k-1}^{(3)}\right|+\left|\sin\theta_{k-1}^{(3)}-\sin\theta_{j-1}^{(2)}\right|=\left(\left|x_i^{(1)}-x_k^{(3)}\right|+\left|\sin\theta_i^{(1)}-\sin\theta_k^{(3)}\right|+\left|\sin\theta_{i-1}^{(1)}-\sin\theta_{k-1}^{(3)}\right|\right)+\left(\left|x_k^{(3)}-x_j^{(2)}\right|+\left|\sin\theta_k^{(3)}-\sin\theta_j^{(2)}\right|+\left|\sin\theta_{k-1}^{(3)}-\sin\theta_{j-1}^{(2)}\right|\right)=d_{BSD}\left(TS_i^{(1)},TS_k^{(3)}\right)+d_{BSD}\left(TS_k^{(3)},TS_j^{(2)}\right)=d_{BSD}(x,z)+d_{BSD}(z,y).$ □

# References

[1] H. Izakian, W. Pedrycz, I. Jamal, Fuzzy clustering of time series data using dynamic time warping distance, Eng. Appl. Artif. Intell. 39 (2015) 235–244. doi:10.1016/j.engappai.2014.12.015.

[2] Y. Sadahiro, T. Kobayashi, Exploratory analysis of time series data: Detection of partial similarities, clustering, and visualization, Comput. Environ. Urban Syst. 45 (2014) 24–33. doi:10.1016/j.compenvurbsys.2014.02.001.

[3] C. Guo, Time Series Clustering Based on ICA for Stock Data Analysis, in: 4th Int. Conf. Wirel. Commun. Netw. Mob. Comput., 2008: pp. 1–4.

[4] B. Chandra, A Multivariate Time Series Clustering Approach for Crime Trends Prediction, in: IEEE Int. Conf. Syst. Man Cybern., 2008: pp. 892–896.

[5] D. Zhou, J. Li, W. Ma, Clustering based on LLE for financial multivariate time series, in: Proc. - Int. Conf. Manag. Serv. Sci. MASS 2009, 2009: pp. 1–4. doi:10.1109/ICMSS.2009.5305089.

[6] V.B. Thinh, D.T. Anh, Time series clustering based on I-k-Means and multi-resolution PLA transform, in: 2012 IEEE RIVF Int. Conf. Comput. Commun. Technol. Res. Innov. Vis. Futur. RIVF 2012, 2012: pp. 0–3. doi:10.1109/rivf.2012.6169835.

[7] J.L. Harvill, N. Ravishanker, B.K. Ray, Bispectral-based methods for clustering time series, Comput. Stat. Data Anal. 64 (2013) 113–131. doi:10.1016/j.csda.2013.03.001.

[8] R. Di Salvo, P. Montalto, G. Nunnari, M. Neri, G. Puglisi, Multivariate time series clustering on geophysical data recorded at Mt. Etna from 1996 to 2003, J. Volcanol. Geotherm. Res. 251 (2013) 65–74. doi:10.1016/j.jvolgeores.2012.02.007.

[9] K.Y. Chan, C.K. Kwong, B.Q. Hu, Market segmentation and ideal point identification for new product design using fuzzy data compression and fuzzy clustering methods, Appl. Soft Comput. 12 (2012) 1371–1378. doi:10.1016/j.asoc.2011.11.026.

[10] S. Aghabozorgi, A. Seyed Shirkhorshidi, T. Ying Wah, Time-series clustering - A decade review, Inf. Syst. 53 (2015) 16–38. doi:10.1016/j.is.2015.04.007.

[11] S. Chandrakala, C.C. Sekhar, A density based method for multivariate time series clustering in kernel feature space, in: Neural Networks, 2008. IJCNN 2008.(IEEE World Congr. Comput. Intell. IEEE Int. Jt. Conf., 2008: pp. 1885–1890.

[12] P. Pereira Rodrigues, J. Gama, J.P. Pedroso, Hierarchical Clustering of Time Series Data Streams, in: IEEE Trans. Knowl. Data Eng., 2007: pp. 1–12. doi:10.1109/TKDE.2007.190727.

[13] K. Kalpakis, D. Gada, V. Puttagunta, Distance measures for effective clustering of ARIMA time-series, in: Proc. 2001 IEEE Int. Conf. Data Min., 2001: pp. 273–280. doi:10.1109/ICDM.2001.989529.

[14] S. Laxman, P.S. Sastry, A survey of temporal data mining, Sadhana. 31 (2006) 173–198.

[15] A. Ahmadi, F. Karray, M.S. Kamel, Flocking based approach for data clustering, Nat. Comput. an Int. J. 9 (2010) 767. doi:10.1007/s11047-009-9173-5.

[16] S. Rani, G. Sikka, Recent Techniques of Clustering of Time Series Data: A Survey, Int. J. Comput. Appl. 52 (2012) 1–9. doi:10.5120/8282-1278.

[17] S. Rodpongpun, V. Niennattrakul, C.A. Ratanamahatana, Selective Subsequence Time Series clustering, Knowledge-Based Syst. 35 (2012) 361–368. doi:10.1016/j.knosys.2012.04.022.

[18] T. Warren Liao, Clustering of time series data - A survey, Pattern Recognit. 38 (2005) 1857–1874. doi:10.1016/j.patcog.2005.01.025.

[19] E.A. Maharaj, P. D'Urso, J. Caiado, Time series clustering and classification, n.d.

[20] F. Yang, T. Sun, C. Zhang, An efficient hybrid data clustering method based on K-harmonic means and Particle Swarm Optimization, Expert Syst. Appl. 36 (2009) 9847–9852. doi:10.1016/j.eswa.2009.02.003.

[21] A.. Ahmadi, F.. Karray, M.S.. Kamel, Model order selection for multiple cooperative swarms clustering using stability analysis, Inf. Sci. (Ny). 182 (2012) 169–183. doi:10.1016/j.ins.2010.10.010.

[22] T. Cura, A particle swarm optimization approach to clustering, Expert Syst. Appl. 39 (2012) 1582–1588. doi:10.1016/j.eswa.2011.07.123.

[23] Y. Kao, E. Zahara, I.-W. Kao, A hybridized approach to data clustering, Expert Syst. Appl. 34 (2008) 1754–1762. doi:10.1016/j.eswa.2007.01.028.

[24] S. Alam, G. Dobbie, Y.S. Koh, P. Riddle, S. Ur Rehman, Research on particle swarm optimization based clustering: A systematic review of literature and techniques, Swarm Evol. Comput. 17 (2014) 1–13. doi:10.1016/j.swevo.2014.02.001.

[25] S. Das, A. Abraham, A. Konar, Automatic kernel clustering with a Multi-Elitist Particle Swarm Optimization Algorithm, Pattern Recognit. Lett. 29 (2008) 688–699. doi:10.1016/j.patrec.2007.12.002.

[26] K.Y. Huang, A hybrid particle swarm optimization approach for clustering and classification of datasets, Knowledge-Based Syst. 24 (2011) 420–426. doi:10.1016/j.knosys.2010.12.003.

[27] A. Ahmadi, F. Karray, M. Kamel, Particle swarm clustering ensemble, in: Proc. 10th Annu. Conf. Genet. Evol. Comput. - GECCO '08, 2008: p. 159. doi:10.1145/1389095.1389118.

[28] L.N. Ferreira, L. Zhao, Time series clustering via community detection in networks, Inf. Sci. (Ny). 326 (2016) 227–242. doi:10.1016/J.INS.2015.07.046.

[29] D. Hallac, S. Vare, S. Boyd, J. Leskovec, Toeplitz Inverse Covariance-Based Clustering of Multivariate Time Series Data, in: Proc. 23rd ACM SIGKDD Int. Conf. Knowl. Discov. Data Min. - KDD '17, ACM Press, New York, New York, USA, 2017: pp. 215–223. doi:10.1145/3097983.3098060.

[30] H. Izakian, W. Pedrycz, I. Jamal, Fuzzy clustering of time series data using dynamic time warping distance, Eng. Appl. Artif. Intell. 39 (2015) 235–244. doi:10.1016/J.ENGAPPAI.2014.12.015.

[31] M. Kozdoba, S. Mannor, Clustering Time Series and the Surprising Robustness of HMMs, (2016). http://arxiv.org/abs/1605.02531 (accessed June 29, 2019).

[32] Y. Rizk, M. Awad, On extreme learning machines in sequential and time series prediction: A non-iterative and approximate training algorithm for recurrent neural networks, Neurocomputing. 325 (2019) 1–19. doi:10.1016/J.NEUCOM.2018.09.012.

[33] H. Kamalzadeh, M. Hahsler, pomdp: Solver for Partially Observable Markov Decision Processes (POMDP). R package version 0.9.1., (2019). https://cran.r-project.org/package=pomdp.

[34] J. Serra, J.L. Arcos, An empirical evaluation of similarity measures for time series classification, Knowledge-Based Syst. 67 (2014) 305–314. doi:10.1016/j.knosys.2014.04.035.

[35] S. Benabderrahmane, T. Guyet, Evaluating Distance Measures and Times Series Clustering for Temporal Patterns, in: IEEE 15th Int. Conf. Inf. Reuse Integr., 2014: pp. 434–441.

[36] E.J. Keogh, M.J. Pazzani, Scaling up dynamic time warping for datamining applications, Knowl. Discov. Data Min. In 6th ACM (2000) 285–289. doi:10.1145/347090.347153.

[37] Y.S. Jeong, M.K. Jeong, O.A. Omitaomu, Weighted dynamic time warping for time series classification, Pattern Recognit. 44 (2011) 2231–2240. doi:10.1016/j.patcog.2010.09.022.

[38] M. Müller, Information Retrieval for Music and Motion, 2007.

[39] E.J. Keogh, M.J. Pazzani, Derivative dynamic time warping, SIAM Conf. Data Min. (2001) 1–11. doi:10.1137/1.9781611972719.1.

[40] Q. Cai, L. Chen, J. Sun, Piecewise statistic approximation based similarity measure for time series, Knowledge-Based Syst. 85 (2015) 181–195. doi:10.1016/j.knosys.2015.05.005.

[41] J. Mei, M. Liu, Y. Wang, H. Gao, Learning a Mahalanobis Distance-Based Dynamic Time Warping Measure for Multivariate Time Series Classification, IEEE Trans. Cybern. 46 (2016) 1363–1374.

[42] T. Górecki, Classification of time series using combination of DTW and LCSS dissimilarity measures, Commun. Stat. - Simul. Comput. 0918 (2017). doi:10.1080/03610918.2017.1280829.

[43] R.J. Kate, Using dynamic time warping distances as features for improved time series classification, Data Min. Knowl. Discov. 30 (2016) 283–312. doi:10.1007/s10618-015-0418-x.

[44] L. Chen, M.T. Özsu, V. Oria, Robust and fast similarity search for moving object trajectories, in: Proc. 2005 ACM SIGMOD Int. Conf. Manag. Data - SIGMOD '05, 2005: pp. 491–503. doi:10.1145/1066157.1066213.

[45] P.-F. Marteau, Time Warp Edit Distance with Stiffness Adjustment for Time Series Matching, IEEE Trans. Pattern Anal. Mach. Intell. 31 (2009) 306–318.

[46] A. Singhal, D.E. Seborg, Clustering of multivariate time-series data.pdf, Chemometrics. 19 (2005) 427–438.

[47] A. Ahmadi, A. Mozafarinia, A. Mohebi, Clustering of Multivariate Time Series Data Using Particle Swarm Optimization, in: Artif. Intell. Signal Process. (AISP), 2015 Int. Symp., Mashhad, 2015. doi:10.1109/AISP.2015.7123516x.

[48] J. Paparrizos, L. Gravano, k-Shape: Efficient and Accurate Clustering of Time Series, in: SIGMOD '15 Proc. ACM SIGMOD Int. Conf. Manag. Data, ACM Press, Melbourne, Victoria, Australia, 2015: pp. 1855–1870. doi:10.1145/2949741.2949758.

[49] J. Paparrizos, L. Gravano, Fast and Accurate Time-Series Clustering, ACM Trans. Database Syst. 42 (2017) 1–49. doi:10.1145/3044711.

[50] F. Höppner, Improving time series similarity measures by integrating preprocessing steps, Data Min. Knowl. Discov. 31 (2017) 851–878. doi:10.1007/s10618-016-0490-x.

[51] M. Gokce, B. George, Time series representation and similarity based on local autopatterns, Data Min. Knowl. Discov. 30 (2016) 476–509. doi:10.1007/s10618-015-0425-y.

[52] H. Kamalzadeh, A. Ahmadi, S. Mansour, A shape-based adaptive segmentation of time-series using particle swarm optimization, Inf. Syst. 67 (2017) 1–18. doi:10.1016/j.is.2017.03.004.

[53] H. Sakoe, S. Chiba, Dynamic Programming Algorithm Optimization for Spoken Word Recognition, IEEE Trans. Acoust. Speech, Signal Process. 26 (1978) 43–49. doi:10.1109/TASSP.1978.1163055.

[54] M. Chiș, S. Banerjee, A.E. Hassanien, Clustering Time Series Data : An Evolutionary Approach, Found. Comput. Intell. 6 (2009) 193–207.

[55] H. Ding, G. Trajcevski, P. Scheuermann, Querying and mining of time series data: experimental comparison of representations and distance measures, Proc. VLDB Endow. 1 (2008) 1542–1552. doi:10.1145/1454159.1454226.

[56] V. Kavitha, M. Punithavalli, Clustering time series data stream-a literature survey, Int. J. Comput. Sci. Inf. Secur. 8 (2010) 289–294. http://arxiv.org/abs/1005.4270.

[57] E. Keogh, S. Kasetty, On the need for time series data mining benchmarks, Data Min. Knowl. Discov. 7 (2003) 349–371. doi:10.1145/775047.775062.

[58] C.M. Antunes, A.L. Oliveira, Temporal Data Mining : an overview, Lect. Notes Comput. Sci.

(2001) 1–15. http://citeseerx.ist.psu.edu/viewdoc/download?doi=10.1.1.97.5516&rep=rep1&type=pdf.

[59] M.Ó. Searcóid, Metric Spaces, in: Numer. Methods Partial Differ. Equ., 2007: pp. 93–104.

[60] C. Cassisi, P. Montalto, M. Aliotta, A. Cannata, A. Pulvirenti, Similarity Measures and Dimensionality Reduction Techniques for Time Series Data Mining, in: Adv. Data Min. Knowl. Discov. Appl., InTech, 2012. doi:10.5772/49941.

[61] J. Kennedy, R. Eberhart, Particle swarm optimization, Neural Networks, 1995. Proceedings., IEEE Int. Conf. 4 (1995) 1942–1948 vol.4. doi:10.1109/ICNN.1995.488968.

[62] E.C. Laskari, K.E. Parsopoulos, M.N. Vrahatis, Particle swarm optimization for integer programming, Proc. 2002 Congr. Evol. Comput. CEC'02. 2 (2002) 1582–1587. doi:10.1109/CEC.2002.1004478.

[63] E. Elbeltagi, T. Hegazy, D. Grierson, Comparison among five evolutionary-based optimization algorithms, Adv. Eng. Informatics. 19 (2005) 43–53. doi:10.1016/J.AEI.2005.01.004.

[64] Y. Shi, R.C. Eberhart, Empirical study of particle swarm optimization, in: Proc. 1999 Congr. Evol. Comput. (Cat. No. 99TH8406), IEEE, n.d.: pp. 1945–1950. doi:10.1109/CEC.1999.785511.

[65] C. Yanping, E. Keogh, B. Hu, N. Begum, A. Bagnall, A. Mueen, G. Batista, UCR Time Series Classification Archive, URL:Www.Cs.Ucr.Edu/~eamonn/Time_series_data/. (2015).

[66] A. Ahmadi, F. Karray, M. Kamel, Multiple cooperating swarms for data clustering, Proc. 2007 IEEE Swarm Intell. Symp. (2007) 206–212.

[67] M. Vlachos, G. Kollios, D. Gunopulos, Discovering similar multidimensional trajectories, in: Proc. 18th Int. Conf. Data Eng., 2002: pp. 673–684. doi:10.1109/ICDE.2002.994784.

[68] G. Taguchi (Asian productivity organization), Introduction to quality engineering: Designing quality into products and processes, Qual. Reliab. Eng. Int. 4 (1986) 198–198. http://doi.wiley.com/10.1002/qre.4680040216.